\documentclass[final]{cvpr}

\usepackage{times}
\usepackage{epsfig}
\usepackage{graphicx}
\usepackage{amsmath}
\usepackage{amssymb}
\usepackage{xcolor}
\usepackage{graphicx}
\usepackage{multirow}
\makeatletter
\robustify\@latex@warning@no@line
\makeatother
\usepackage{authblk}

\usepackage[pagebackref=true,breaklinks=true,colorlinks,bookmarks=false]{hyperref}
\def\cvprPaperID{30} % *** Enter the CVPR Paper ID here
\def\confYear{CVPR 2021}

\makeatletter
\newcommand{\printfnsymbol}[1]{%
  \textsuperscript{\@fnsymbol{#1}}%
}

\begin{document}

\title{Real-time Monocular Depth Estimation with Sparse Supervision on Mobile}

\author[1]{Mehmet Kerim Yucel}
\author[2]{Valia Dimaridou}
\author[2]{Anastasios Drosou}
\author[1]{Albert Sa\`a-Garriga}
\affil[1]{Samsung Research UK}
\affil[2]{Centre for Research and Technology Hellas (CERTH), Information Technologies Institute, Thessaloniki, Greece}
\affil[ ]{\textit {\{mehmet.yucel, a.garriga\}@samsung.com, {\{valia, drosou\}@iti.gr}}}

\maketitle

%%%%%%%%% ABSTRACT
\begin{abstract}
%\begin{itemize}
Monocular (relative or metric) depth estimation is a critical task for various applications, such as autonomous vehicles, augmented reality and image editing. In recent years, with the increasing availability of mobile devices, accurate and mobile-friendly depth models have gained importance. Increasingly accurate models typically require more computational resources, which inhibits the use of such models on mobile devices. The mobile use case is arguably the most unrestricted one, which requires highly accurate yet mobile-friendly architectures. Therefore, we try to answer the following question: \textbf{How can we improve a model without adding further complexity (i.e. parameters)?}

Towards this end, we systematically explore the design space of a relative depth estimation model from various dimensions and we show, with key design choices and ablation studies, even an existing architecture can reach highly competitive performance to the state of the art, with a fraction of the complexity. Our study spans an in-depth backbone model selection process, knowledge distillation, intermediate predictions, model pruning and loss rebalancing. We show that our model, using only DIW as the supervisory dataset, achieves 0.1156 WHDR on DIW with 2.6M parameters and reaches 37 FPS on a mobile GPU, without pruning or hardware-specific optimization. A pruned version of our model achieves 0.1208 WHDR on DIW with 1M parameters and reaches 44 FPS on a mobile GPU.

\end{abstract}

\section{Introduction}
\label{sec:intro}

The acquisition of accurate depth information from a scene is an integral part of computer vision, as it provides crucial information of the present 3D structure. This information is imperative to various applications, such as augmented reality, compositing, scene manipulation and robotics.  Accurate depth information has traditionally been acquired using multi-camera/stereo setups, LIDARs and other specialized sensors. The use of such sensors in mobile/edge devices may dramatically increase the cost or may not be feasible due to other constraints. Depth estimation using a single camera offers a simpler and a low-cost alternative to such traditional setups.

\begin{figure}[]%
\begin{center}
\vspace{9mm}
\includegraphics[width=\columnwidth]{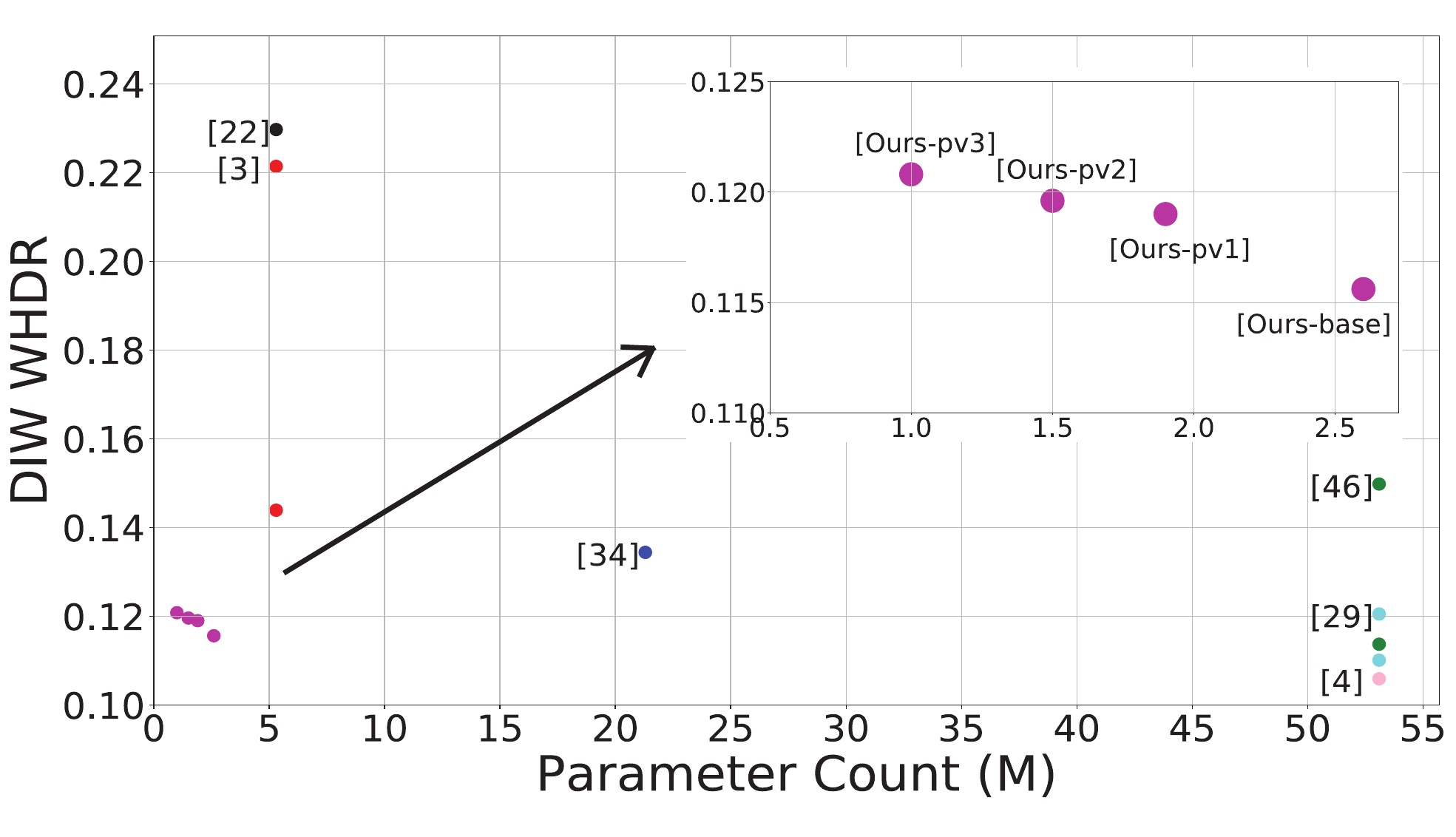}
\end{center}
\vspace{-3mm}
    \caption{Parameter count vs. DIW WHDR scores. Our models achieve highly competitive results, despite being significantly smaller. Each color represents one method; duplicates of each color represent different results using the same architecture (except the ones denoted by \textit{ours}). Refer to Table \ref{tab:sota_comparison} for more details. }
\label{method:page1}
\vspace{-5mm}
\end{figure}

Learning-based methods which model the necessary cues for depth estimation have been proposed in \cite{eigen2014depth,eigen2015predicting, liu2015deep, shelhamer2015scene, laina2016deeper}. Parallel to these works, instead of predicting metric depth, in-the-wild depth estimation scenarios opted to predict relative depth estimates  \cite{xian2020structure, xian2018monocular, chen2016single}. Self-supervised methods were also shown to be viable alternatives for monocular depth estimation \cite{garg2016unsupervised, godard2019digging, johnston2020self}. 

The literature primarily focuses on higher accuracy, at the cost of runtime performance and compute requirements, which are not feasible for mobile/edge applications. To address this, lightweight depth estimators have been proposed \cite{wofk2019fastdepth}, either by using small backbones or by systematically designing efficient depth estimators. However, such designs have generally been evaluated on restricted cases and not in-the-wild scenarios, which mobile/edge devices operate in.

In this paper, we perform a systematic study and present a detailed pipeline to generate a compact monocular relative depth estimation model.  We evaluate and  adapt, with key design choices, existing advances in the field, such as knowledge distillation \cite{liu2020structured}, intermediate depth predictions \cite{godard2019digging}, loss rebalancing \cite{lee2020multi} and pruning \cite{yang2018netadapt}. The contributions of our paper are summarized as follows:

\begin{itemize}
\item We alter a knowledge distillation pipeline \cite{liu2020structured} and show that a pixel-wise regression loss, with a suitable teacher network, achieves higher accuracy.

\item We augment a loss rebalancing pipeline \cite{lee2020multi} to work effectively with intermediate prediction layers and show that balancing the decay rate of loss terms and emphasising distillation losses in early training stages help achieve accuracy boosts.

\item Our model, despite being 20x smaller than the state-of-the-art and using only DIW as the supervisory dataset, achieves 0.1156 WHDR on DIW  and runs at 37 FPS on a mobile GPU. A pruned version of our model, despite being 50x smaller than the state-of-the-art, achieves 0.1208 WHDR and runs at 44 FPS on a mobile GPU.

\end{itemize}

\section{Related Work} \label{related_work}

\subsection{Monocular Depth Estimation}
Monocular depth estimation is inherently an ill-posed problem as there is no unique solution (i.e. The same scenes can be projected to the 2D space using non-unique depth maps). Despite this, several cues can be used to restrict the solution space. Using the combination of local predictions \cite{saxena2008make3d}, semantic-guided predictions \cite{liu2010single}, non-parametric sampling \cite{karsch2014depth}, retrieval \cite{konrad20122d} and super-pixel  optimization \cite{liu2014discrete} methods are examples of such approaches. 

Seminal work of Eigen et al. \cite{eigen2014depth} showed a two-stage CNN can learn to infer depth without explicit feature crafting, paving the way for end-to-end solutions for depth estimation. Laina et al. \cite{laina2016deeper} utilized an altered version of ResNet-50 and showed accuracy improvements. Lee et al. \cite{lee2018single} showed frequency-domain aggregation of depth map candidates can produce accurate depth maps. Hu et al. \cite{hu2019revisiting} showed a feature fusion scheme can improve predictions. Intermediate depth predictions \cite{godard2019digging}, loss rebalancing \cite{lee2020multi}, semantics-driven depth predictions \cite{wang2020sdc}, depth estimation with uncertainty \cite{poggi2020uncertainty}, attention \cite{xu2018structured} and geometric constraints \cite{yin2019enforcing} have also proven to be viable improvements.

Such advances have been made possible thanks to large datasets with dense depth ground-truths, such as NYUv2 \cite{Silberman:ECCV12}, KITTI \cite{eigen2015predicting}, Mannequin Challenge (MC) \cite{li2019learning} and others. Metric depth annotations are, however, quite laborious to obtain, suffer from scale incompabilities and can fail to generalize to unconstrained settings. These problems are partially alleviated by self-supervision, where minimizing a form of reconstruction error using unlabeled monocular sequences is shown to be a viable alternative \cite{garg2016unsupervised, godard2019digging, johnston2020self}. A parallel line of work focuses on relative depth estimation, especially for in-the-wild scenarios, by using ordinal ground-truths between pairs of pixels. It was shown that relative depth ground-truth can be used to infer dense or relative depth maps successfully \cite{chen2016single}. Several relative depth datasets are proposed \cite{xian2018monocular, xian2020structure} to facilitate research in this area.

\subsection{Lightweight Models}

The advances in depth estimation generally come at the cost of increased computational budgets. For mobile/edge applications, it is imperative to have an accurate model that can perform in real-time in resource-constrained environments. A natural starting point is to replace the backbones of existing models with lightweight alternatives, such as MobileNet \cite{sandler2018mobilenetv2}, GhostNet \cite{han2020ghostnet} and Fbnet \cite{wu2019fbnet}. Additionally, knowledge distillation \cite{liu2020structured}, reduced precision training and  network pruning \cite{yang2018netadapt} are common approaches to reduce model size. Several other studies focused on producing efficient depth estimators from scratch, such as feature-pyramid based models \cite{poggi2018towards}, low-latency decoder designs \cite{wofk2019fastdepth} and reinforcement-learning based pruning models \cite{tu2020efficient}.

A similar work to ours is reported in \cite{aleotti2021real}. Authors compare several lightweight depth estimation models by using a knowledge-distillation based training and perform cross-dataset experiments to assess generalization. Our study differs in several aspects: \textbf{(1)} we present a detailed evaluation and design process for architecture selection, \textbf{(2)} show that a combination of knowledge distillation, \textbf{(3)} intermediate prediction layers and \textbf{(4)} loss rebalancing can achieve  competitive results using relative depth ground-truth.

\section{Methodology} 
\label{sec:method}
In this section, we outline our design choices and systematic study to produce an efficient depth estimator.

\subsection{Model Design} \label{sec:model_design}
One of the most critical parts of a machine-learning process is the design of the architecture. We base our model design on U-Net \cite{ronneberger2015u} and systematically improve the design by  selecting encoder and decoder topologies.

First, we evaluate several encoders while keeping the decoder fixed (except matching the required filter numbers of the encoder) as the one in \cite{wofk2019fastdepth}. We evaluate ShuffleNetv2 \cite{ma2018shufflenet}, MNasNet \cite{tan2019mnasnet}, MobileNetv2 \cite{sandler2018mobilenetv2}, MobileNetv3 \cite{howard2019searching}, EfficientNet \cite{tan2019efficientnet} variants, MixNet \cite{tan2019mixconv}, GhostNet \cite{han2020ghostnet} and FastDepth \cite{wofk2019fastdepth}. Second, we focus on the decoder design. For the various encoders mentioned above, we compare NNConv5 \cite{wofk2019fastdepth} and an FbNet-based \cite{wu2019fbnet} decoder layers. 

Our final architecture is based on a MobileNetv2 encoder and an FbNet-based decoder. We observe MobileNetv2 to have a good trade-off between efficiency and accuracy, whereas the FbNet based decoder is smaller and more accurate than the alternatives. We also update the decoder by replacing its last layer with a simple upsampling layer and include five skip connections from encoder to decoder. Our final network architecture is shown in Figure \ref{methodFig}.

\subsection{Motivation and Baselines} \label{sec:baselines}
\subsubsection{Motivation} \label{sec:motivation}

\textbf{Model performance.} First, we need to consider the use of mobile depth estimators; they need to work well in arguably the most in-the-wild setting. Increasing the model complexity for a mobile device is generally not an option. Therefore, we need to answer the first question; \textit{How can we improve a model's performance without adding additional complexity?} We try to answer this question in Section \ref{section:ours}

\textbf{Ground-truth.} Second, we need to consider the nature of ground-truth; one may  assume metric depth annotations are required for in-the-wild settings. However, depending on the use case, metric depth may not be necessary. In applications such as image relighting and visual effects, we may not need to know the metric depth for each pixel, but knowing their ordering relative to other pixels can be sufficient. Therefore, we need to answer the second question; \textit{What type of ground-truth annotations do we want?} 

Ordinal ground truths, where point-pair relations for one or several points per image are given, are suitable annotations for such scenarios. Relative depth annotations are easier to generate, do not suffer from scale incompatibilities and are more robust to outliers \cite{chen2019learning}. Moreover, relative depth models can learn strong priors, which can then be projected to any scale in metric depth scenarios. In light of these, we opt to use relative ground-truth annotations.
\vspace{-3mm}
\subsubsection{Baselines} \label{subsec:baselines}

We select datasets that best reflect the in-the-wild case; RedWeb \cite{xian2018monocular}, DIW \cite{chen2016single} and Mannequin Challenge \cite{li2019learning}. The first two have relative ground-truths, which are suitable for our use case. We base our study on performance on DIW, due to several reasons; i) it represents in-the-wild the best due to its size and variance, ii) it is arguably the most suitable datasets for relative depth and iii) its evaluation metrics are the widely used standard for in-the-wild scenarios. Evaluating our improvements requires a baseline, and towards this end, we train our model on each dataset and also sequentially on all of them to serve as a baseline. Next, we formulate the losses used to train our model.

\begin{figure*}[!ht]%
\begin{center}
\vspace{-6mm}
\includegraphics[width=\textwidth]{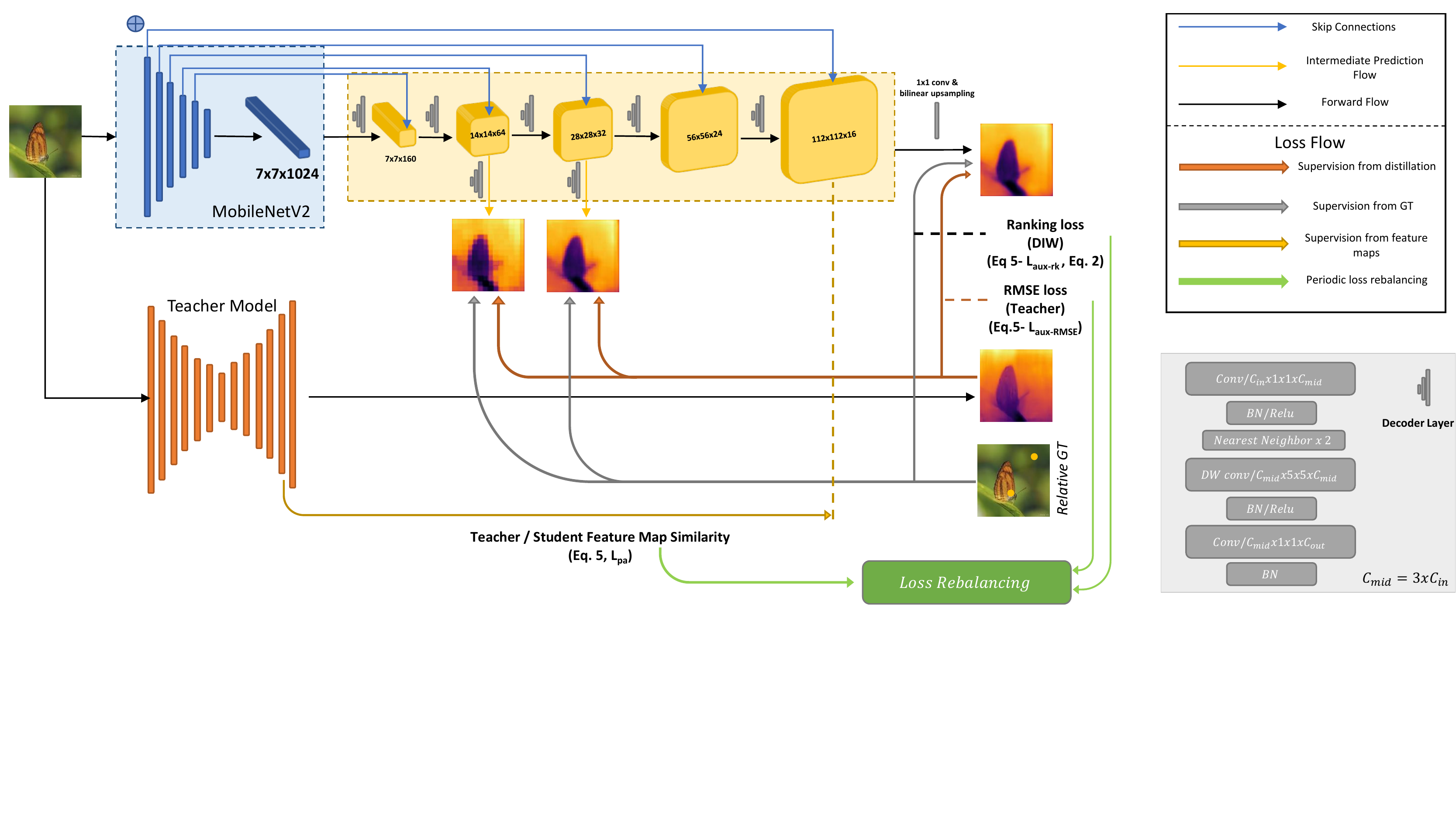}
\end{center}
\vspace{-2cm}
    \caption{Our architecture and training pipeline. The decoder layers based on FBNet \cite{wu2019fbnet} are shown in detail in bottom right. We train our model on DIW ($L_{aux-rk}$ in Eq. (\ref{eq:kd})) and also distill knowledge \cite{liu2020structured} from a high-capacity model \cite{chen2019learning} ($L_{pa}$ and $L_{aux-RMSE}$ in Eq. (\ref{eq:kd})). During training, we train with intermediate prediction layers which are removed during inference. Our multiple loss terms are balanced during training in real-time \cite{lee2020multi}. }
\label{methodFig}
\vspace{-5mm}
\end{figure*}

\textbf{Mannequin Challenge.} For training our model on Mannequin Challenge dataset, we use the losses formulated in the original paper \cite{li2019learning}. The final loss function consists of three loss terms, which is defined as 

\begin{equation} \label{eq:mc}
L_{si} = L_{mse} + a_{1}L_{grad} + a_2L_{sm}
\end{equation}

\noindent where the first term $L_{mse}$ is the scale-invariant mean squared error, the second term $L_{grad}$ is multi-scale gradient term that encourages smoother gradient changes and sharper depth discontinuities in the predictions, the third term $L_{sm}$ is the multi-scale edge-aware smoothness term that encourages smooth interpolation of depth values in textureless regions and $a_1/a_2$ are loss weights.

\textbf{DIW.} Training on DIW is based on the ranking loss \cite{chen2016single}, which is defined as

\begin{equation}
    \psi_{k}(I, i_k, j_k, r_k,z)= 
\begin{cases}
    r_k=1, & \log(1 + exp(z_{jk}  - z_{ik})) \\
    r_k=-1,&  \log(1 + exp(z_{ik}   - z_{jk})) \\
    r_k=0, & (z_{ik} - z_{jk})^{2}
\end{cases} \label{eq:diw}
\end{equation}

\noindent where  $i_k$, $j_k$ and $r_k$ represent first point, second point and their ground-truth relations for the query $k$ in training image $I$ and $z$ is the predicted depth map. Ground-truth relations are encoded as ${1, -1, 0}$ for ordinal relations \textit{closer, further} and \textit{equal}, respectively.

\textbf{RedWeb.} Training on RedWeb is done with an improved ranking loss \cite{xian2018monocular}, which is defined as\footnote{Readers are referred to original papers for further details.} 
 
\begin{equation} \label{eq:rw}
    \psi_{k}= 
    \begin{cases}
    l_{k} \neq 0, & \log(1+exp((-z_{ik} + z_{jk}) l_k)) \\
    l_{k}=0 ,&  (z_{ik} - z_{jk})^{2}
\end{cases}
\end{equation}
 
\noindent where $\psi$ is parametrized by input image $I$, $z$ is the estimated depth map, and $i_j, j_k, l_k$ represent first point, second point and their ordinal relations, respectively.

\subsection{Towards a Better Depth Estimator} \label{section:ours}

Now that we have finalized our model design in Section \ref{sec:model_design} and obtained baselines in Section \ref{sec:baselines}, we explore ways to improve our model's accuracy, without adding additional parameters or compute requirements.

\textbf{Intermediate Predictions.} First, we integrate intermediate prediction layers to our model \cite{godard2019digging,zhang2019your}, which produce multiple outputs during training. One key difference here is that we remove the intermediate prediction layers after training, and thus preserve the model complexity. The use of  intermediate predictions in decoder layers acts as a regularizer and it forces the encoder to learn better representations. We train our network on the combined loss of last and intermediate prediction layers, which is defined as 

\begin{equation} \label{eq:aux}
L_{aux} =  \sum_{i=1}^{N}{ \lambda_i L_{i}} 
\end{equation}

\noindent where $L_i$ is the loss\footnote{$L_i$ is the ranking loss (Equation (\ref{eq:diw})) in second term of Equation (\ref{eq:kd}) and RMSE in the first term of Equation (\ref{eq:kd}).}, $\lambda_i$ is the weight term and $i{\in}N$ represents the (intermediate and final) predictions\footnote{In our equations, the summation operator over batches is omitted for brevity.}. Our final model has two intermediate prediction layers based on FbNet-like upsampling layers that are used in our decoder. We integrate two intermediate prediction layers after second and third decoder layers and use loss weights $0.5, 0.25, 0.25$ for the final and intermediate predictions, respectively.

\textbf{Knowledge Distillation.}
Second, we use a knowledge distillation mechanism to exploit knowledge learned by larger models. Knowledge distillation is data-efficient method to transfer the knowledge of a large model to a compact model, and leads to accuracy improvements without adding further complexity to the model.

Our knowledge distillation mechanism follows the work of Liu et al. \cite{liu2020structured}, where we use as teacher the EncDecResNet architecture trained on RedWeb, DIW and Y3D datasets \cite{chen2019learning}. The original knowledge distillation approach formulates three losses to transfer the knowledge to the student; a pixel-wise loss operating on binned depth classes (between student and teacher networks), pair-wise loss operating on feature representations of teacher and student network, and a holistic loss that also trains a discriminator such that the student (i.e. generator) outputs accurate depth maps.

Our implementation differs from the original in several aspects; i) we change the classification-based formulation with simple RMSE for pixel-wise loss, ii) remove holistic loss and iii) use a different teacher network. We select our teacher based on its performance on DIW. We use RMSE for pixel-wise loss since no scale issues are expected (teacher is trained on DIW) and RMSE is one of the strongest supervision available in this scenario. We remove holistic loss since we observe our model reaches its capacity rather quickly, and generator/discriminator pair fail to converge in this time frame. Finally, we heuristically select the feature layers for pair-wise loss and set our student model as the model explained in previous sections, with the additional intermediate prediction layers. Our model is trained with the loss defined as

\begin{equation} \label{eq:kd}
\begin{split}
L_{kd} =  \lambda_1 L_{aux^{RMSE}}(S(I), \tau(I)) + \lambda_2 L_{aux^{rk}} (S(I), Y(I)) \\ 
+ \lambda_3 L_{pa}(S(I)_{\ell}, \tau(I)_{\ell})
\end{split}
\end{equation}

\noindent where $S$ and  $\tau$ are student and teacher networks, $I$ is the training sample, $Y(.)$ is the ground-truth label,  $L_{pa}$ is the pairwise loss operating on feature representations $\ell$, and $\lambda$ values are loss weights. We enforce pixelwise  ($RMSE$) and ranking loss ($rk$) to every prediction, and enforce pairwise loss between penultimate (decoder) layers of $S$ and  $\tau$.

\textbf{Loss Rebalancing.} As can be seen in Equations (\ref{eq:aux}) and (\ref{eq:kd}), there are multiple loss terms driving the model training, which results in another layer of parameter tuning. This is a laborious task as loss values have different value scales and decay rates, which requires periodic scaling of loss weights to facilitate adequate contribution for each loss term.

We alleviate this problem by using automatic loss balancing algorithm proposed in \cite{lee2020multi}. Authors of the original paper formulate an algorithm that i) performs an initial training stage and records the loss values, ii) scales the values to have a level playing field across loss terms and iii) periodically updates the weights by taking into account the rate of decrease for each loss. The algorithm is a scalable one as it also allows emphasising \textit{hard} or \textit{easy} losses first, which are defined so based on the way they decay.

We integrate the said loss rebalancing algorithm with key differences; i) we do not include separate weights for auxiliary layers and keep them wrapped (i.e. $L_{aux}$ shown in Equation (\ref{eq:aux}) has fixed $\lambda$ values), ii) we omit the loss weight initialization/scaling stage and set all weights equal to each other iii) emphasize hard losses first and then easy losses during our training. We do not include separate weights for auxiliary layers because doing so means giving more weights to intermediate losses, which can lead to insufficient supervision for the deeper layers of the decoder. We omit the initial loss weight scaling because we observe that different loss terms have significantly different value scales, and levelling these value scales earlier in the training is likely to degrade the performance. Moreover, we see that focusing on the hard losses first emphasise distillation losses in early training, which essentially is a form of pretraining. We hypothesize this setting is likely to be more beneficial. Essentially, instead of manually choosing $\lambda_1$, $\lambda_2$ and $\lambda_3$ of Equation (\ref{eq:kd}), loss rebalancing adaptively chooses and changes them during training.

\textbf{Final Words.} The final training pipeline is shown in Figure \ref{methodFig}. We also prune our model using NetAdapt \cite{yang2018netadapt}; we remove the intermediate prediction layers, prune the model and finetune on DIW using the same training pipeline.

\section{Experiments} \label{results}

\subsection{Experimental Setup}

We use PyTorch \cite{paszke2019pytorch} for training and convert the models to TFLite for performance evaluation \cite{abadi2016tensorflow}. We load ImageNet-pretrained weights for our encoder and initialize the weights using \cite{he2015delving} for the decoder. 

\textbf{NYUv2 Depth.} \cite{Silberman:ECCV12} includes 48K RGB-D images with a size of 640x480 and the majority of images are in indoor settings. Similar to \cite{wofk2019fastdepth}, the model is trained with a batch size of 8 and a starting learning rate of 0.01, which decays by an order of magnitude every 10 epochs. We train for a total of 30 epochs and optimize with SGD, with a momentum of 0.9 and a weight decay of 0.0001. The training is performed with $\ell_1$ loss. We use scaling, rotation, color jitter/normalization and random flipping as training augmentations.  Augmented data is then resized and center-cropped to 224x224. We use RMSE and  $\delta_1$ metrics for evaluation. NYUv2 dataset is used in Section \ref{sec:model_design}, where the model designs are compared for various encoder/decoder pairs.

\textbf{Mannequin Challenge.} includes 2695 monocular sequences ($\sim$150K frames) with a size of 640x480. Despite its size, since the dataset primarily includes people, its scene distribution is still restricted. The model is trained with a batch size of 16 and a starting learning rate of 0.0004, using the loss function shown in Equation (\ref{eq:mc}). Training is done for 12 epochs with Adam \cite{kingma2014adam} optimiser, where we halve the learning rate every 4th epoch. We use three scales for $L_{sm}$ in Equation (\ref{eq:mc}) and use rotation, color and depth normalization and random flipping as data augmentations. The data is finally resized and then random-cropped to 224x224. We use Scale-Invariant RMSE as the evaluation metric \cite{li2019learning}. MC dataset is used in Section \ref{sec:baselines} to create our baselines.

\textbf{RedWeb.} includes 3600 images with dense and relative depth annotations for every pixel in the image \cite{xian2018monocular}. Despite its large scene variance, its size is a limiting factor. The model is trained using a batch size of 4 and a starting learning rate of 0.0004, with the loss function shown in Equation (\ref{eq:rw}). Training is done for 250 epochs with Adam optimizer, where learning rate is halved every 50th epoch. We sample 3000 points for each image randomly and augment the data with rotation, color jitter, color normalization and random flips. The data is resized and random-cropped to 224x224. RedWeb is used in Section \ref{sec:baselines} to create our baselines.

\textbf{DIW.} includes 495K images with relative depth annotations, where a point-pair is sampled per image. DIW is the largest among all others, both in size and distribution. We train using a batch size of 4 and a starting learning rate of 0.0001, with the loss function shown in Equation (\ref{eq:diw})\footnote{We add other losses to DIW training in Section \ref{section:ours}.}. Training is performed for 5 epochs and Adam optimizer is used, where we halve the learning rate every epoch. The images are resized to 224x448 resolution and then fed to the network. We use WHDR as the evaluation metric \cite{chen2016single}. DIW is used in Section \ref{sec:baselines} to create our baselines, as well as in Section \ref{section:ours} to train our final depth estimator\footnote{RMSE, WHDR and SI-RMSE are measured on NYUv2, DIW and MC datasets, respectively in Tables \ref{tab:baseline} to \ref{tab:ablation_lossrebalancing}.}. 

\subsection{Model Design}
Results for encoder architectures are shown in Table \ref{tab:model_design_1}, where EfficientNet-based encoders have the best accuracy. In overall, however, MobileNetv2 has the best complexity-accuracy trade-off, therefore it is the best choice available.

\begingroup
\setlength{\tabcolsep}{8pt} % Default value: 6pt
\renewcommand{\arraystretch}{1.5} % Default value: 1
\begin{table}[]
\resizebox{\columnwidth}{!}{%
\begin{tabular}{lccc}
\hline
\textbf{Model} & \textbf{RMSE} $\downarrow$ & \textbf{$\delta_1$} $\uparrow$ & \textbf{Parameters (M)} \\ \hline

ShuffleNet v2 \cite{ma2018shufflenet} & 0.615 & 0.749 & 2.0 \\ \hline
MNasNet \cite{tan2019mnasnet} & 0.608 & 0.758 & 4.0 \\ \hline
MobileNet v2 \cite{sandler2018mobilenetv2} & 0.583 & 0.775 & 3.1 \\ \hline
MobileNet v3 \cite{howard2019searching} & 0.607 & 0.749 & 5.1 \\ \hline
EfficientNet ES \cite{tu2020efficient} & 0.572 & 0.782 & 5.0 \\ \hline
EfficientNet B0 \cite{tu2020efficient} & 0.581 & 0.778 & 4.9 \\ \hline
GhostNet \cite{han2020ghostnet} & 0.618 & 0.750 & 5.4 \\ \hline
MixNet M \cite{tan2019mixconv} & 0.589 & 0.767 & 4.5 \\ \hline
MixNet S \cite{tan2019mixconv} & 0.589 & 0.766 & 3.6 \\ \hline
FastDepth \cite{wofk2019fastdepth} & 0.599 & 0.775 & 3.9 \\ \hline
\end{tabular}%
}
\medskip
\caption{Different encoder architectures' performance on NYUv2, where decoder is fixed to NNConv5 \cite{wofk2019fastdepth}.}
\vspace{-3mm} 
\label{tab:model_design_1}
\end{table}
\endgroup

Results for decoder architecture comparisons are shown in Table \ref{tab:model_design_2}. There is a consistent trend regardless of the encoders; FBNet-based decoders outperform NNConv5 decoders while having significantly fewer parameters. Seeing that \textit{MNet v2 + FBNet} model gives the best parameter/accuracy trade-off, we further remove its last layer of the decoder and replace it with a simple upsampling operation (last row, Table \ref{tab:model_design_2}) and see that it outperforms the original architecture (penultimate row, Table \ref{tab:model_design_2}). Our architecture, therefore, is chosen as \textit{MNet v2 + FBNetx112} and will be the one used in the following sections.

\begingroup
\setlength{\tabcolsep}{8pt} % Default value: 6pt
\renewcommand{\arraystretch}{1.5} % Default value: 1
\begin{table}[]
\resizebox{\columnwidth}{!}{%
\begin{tabular}{lccc}
\hline
\textbf{Model} & \textbf{RMSE} $\downarrow$ & \textbf{$\delta_1$} $\uparrow$ & \textbf{Parameters (M)} \\ \hline

EfNet ES \cite{tu2020efficient} + NNConv5 \cite{wofk2019fastdepth} & 0.572 & 0.782 & 5.0 \\ \hline
EfNet ES \cite{tu2020efficient} + FBNet \cite{wu2019fbnet} & 0.534 & 0.802 & 4.7 \\ \hline
MixNet M \cite{tan2019mixconv} + NNConv5 \cite{wofk2019fastdepth} & 0.589 & 0.766 & 4.5 \\ \hline
MixNet M  \cite{tan2019mixconv} + FBNet \cite{wu2019fbnet} & 0.582 & 0.773 & 3.6 \\ \hline
GhostNet \cite{han2020ghostnet} + NNConv5 \cite{wofk2019fastdepth} &
0.618 & 0.750 & 5.4 \\ \hline
GhostNet \cite{han2020ghostnet} + FBNet \cite{wu2019fbnet} & 0.596 & 0.756 & 4.3 \\ \hline
MNet v2 \cite{sandler2018mobilenetv2} + NNConv5 \cite{wofk2019fastdepth} & 0.583 & 0.775 & 3.1 \\ \hline
MNet v2 \cite{sandler2018mobilenetv2} + FBNet \cite{wu2019fbnet} & 0.567 & 0.782 & 2.6 \\ \hline
MNet v2 \cite{sandler2018mobilenetv2} + FBNet \cite{wu2019fbnet} x112 & 0.564 & 0.790 & 2.6 \\ \hline
\end{tabular}%
}
\medskip
\caption{Different decoder architectures' performance on NYUv2. Our model is shown in the last row, which has the best performance/accuracy trade-off. }
\vspace{-5mm} 
\label{tab:model_design_2}
\end{table}
\endgroup

\subsection{Baselines}
Our baseline results are shown in Table \ref{tab:baseline}. For NYUv2 and MC, best results are obtained when we train on them, as expected. The second row shows that we achieve 0.1484 WHDR in DIW, if we train on DIW exclusively. When we train on MC, RedWeb and DIW sequentially, we get significant improvements and achieve 0.1316 WHDR. This shows that the distribution of DIW is large and can make use of other datasets with comparably limited data distribution. Our baseline is shown in the last row of Table \ref{tab:baseline}, which is an improved version of the \textit{vanilla} training on DIW.

\begingroup
\setlength{\tabcolsep}{8pt} % Default value: 6pt
\renewcommand{\arraystretch}{1.5} % Default value: 1
\begin{table}[]
\resizebox{\columnwidth}{!}{%
\begin{tabular}{lccc}
\hline
\textbf{Training Dataset} & \textbf{RMSE $\downarrow$} & \textbf{WHDR $\downarrow$} & \textbf{SI-RMSE $\downarrow$} \\ \hline
NYUv2 & 0.564 & 0.3294 & 0.3809 \\ \hline
DIW & 1.332 & 0.1484 & 0.4662 \\ \hline
RedWeb & 1.326 & 0.2046 & 0.3973 \\ \hline
MC & 1.274 & 0.2401 & 0.1097 \\ \hline
RedWeb $\rightarrow$ DIW & 1.319 & 0.1386 & 0.4999 \\ \hline
MC $\rightarrow$ DIW & 1.326 & 0.1376 & 0.3973 \\ \hline
MC $\rightarrow$ RedWeb $\rightarrow$ DIW & 1.313 & 0.1316 & 0.4551 \\ \hline
\end{tabular}%
}
\medskip
\caption{Our model trained on various datasets. $\rightarrow$ indicates sequential training. The last row shows our baseline.}
\vspace{-4mm} 
\label{tab:baseline}
\end{table}
\endgroup

\subsubsection{Ablation Study} \label{sec:ablation_study}
We now study the contributions of our design choices; intermediate predictions, knowledge distillation and loss rebalancing. In this section, we perform training \textit{only on DIW}, but evaluate on all three datasets (MC, RedWeb, DIW).

\textbf{Intermediate Predictions.} We first study the effect of having intermediate prediction layers in our model. We conduct experiments in two training settings to assess intermediate predictions' usefulness; we train on all three datasets sequentially  and also train on DIW with knowledge distillation. For both settings, we show results with one and two intermediate prediction layers, where loss weights are $\lambda_1=0.5, \lambda_2=0.5$  and $\lambda_1=0.5, \lambda_2=\lambda_3=0.25$ for one and two intermediate prediction layers, respectively\footnote{We experiment with different numbers of intermediate prediction layers and loss weights, but we do not include all the experiments for brevity.}. Results are shown in Table \ref{tab:ablation_aux}.

Results show intermediate prediction layers achieve better results across different datasets. Adding one intermediate prediction layer has a degrading effect, but when we add two intermediate prediction layers we get improvements. In sequential training on all our datasets, we get slight loss in WHDR, but achieve improvements in SI-RMSE and RMSE. When trained on DIW with knowledge distillation, we get modest improvements with intermediate prediction on SI-RMSE and WHDR, but lose some accuracy in NYU. In overall, intermediate predictions help achieve better accuracy, especially when we train on DIW with knowledge distillation,  without additional model complexity.

\textbf{Knowledge distillation}
We now study the effect of knowledge distillation\footnote{In Tables \ref{tab:ablation_kd} to \ref{tab:sota_comparison}, \textit{KD} stands for supervision from the teacher.} pipeline; we analyse the contribution of each loss we introduce to the training, and compare our results to the baseline we created in Section \ref{sec:baselines}. We set equal loss weights $\lambda_1=\lambda_2=1.0$ for pixelwise and pairwise losses (refer to Equation (\ref{eq:kd})) throughout our experiments and set intermediate prediction layer loss weights as $\lambda_1=0.5, \lambda_2=\lambda_3=0.25$ (refer to Equation (\ref{eq:aux})). We also experiment with the holistic loss term defined in the original paper; we use hinge loss for adversarial training and set the loss weight as 0.05. We update the discriminator every two iterations to inhibit the learning process of the discriminator. Results are shown in Table \ref{tab:ablation_kd}.

\begingroup
\setlength{\tabcolsep}{8pt} % Default value: 6pt
\renewcommand{\arraystretch}{1.5} % Default value: 1
\begin{table}[]
\resizebox{\columnwidth}{!}{%
\begin{tabular}{lccc}
\hline
\textbf{Model and Training Data} & \textbf{RMSE $\downarrow$} & \textbf{WHDR $\downarrow$} & \textbf{SI-RMSE $\downarrow$} \\ \hline
%No IP (DIW)  & N/A & 0.1484 & N/A \\
%1 IP (DIW)  & N/A & \textbf{0.1354} & N/A \\ \hline
No IP  (MC $\rightarrow$ RedWeb $\rightarrow$ DIW) & 1.313 & \textbf{0.1316} & 0.4551 \\
1 IP  (MC $\rightarrow$ RedWeb $\rightarrow$ DIW) & 1.312 & 0.1331 & \textbf{0.4179} \\
2 IP (MC $\rightarrow$ RedWeb $\rightarrow$ DIW)  & \textbf{1.308} & 0.1337 & 0.4441 \\ \hline
No IP (DIW + knowledge distillation) & \textbf{1.294} & 0.1202 & 0.4396 \\
1 IP (DIW + knowledge distillation) & 1.298 & 0.1214 & 0.4467 \\
2 IP (DIW + knowledge distillation) & 1.303 & \textbf{0.1198} & \textbf{0.4213} \\ \hline
\end{tabular}%
}
\medskip
\caption{Contributions of intermediate predictions (IP). Number of intermediate predictions are shown in model details.}
\vspace{-3mm} 
\label{tab:ablation_aux}
\end{table}
\endgroup

Results show that knowledge distillation introduces significant improvements over our baseline, despite using only DIW as the supervisory dataset. Moreover, including only the pixel-wise loss improves our results, which suggests that although the teacher network is trained on relative depth annotations, using metric supervision (via RMSE) from the teacher is a viable way to distill knowledge, both in terms of relative and metric depth accuracy. Pairwise loss does introduce slight improvements, however, holistic loss fails to introduce any improvements. This is due to the fact that our model overfits after five epochs, and the discriminator fails to converge in this limited training duration. The final model with two intermediate predictions produces the best scores, validating our  design process.

\begingroup
\setlength{\tabcolsep}{8pt} % Default value: 6pt
\renewcommand{\arraystretch}{1.5} % Default value: 1
\begin{table}[]
\resizebox{\columnwidth}{!}{%
\begin{tabular}{lccc}
\hline
\textbf{Model and Training Data} & \textbf{RMSE $\downarrow$} & \textbf{WHDR $\downarrow$} & \textbf{SI-RMSE $\downarrow$} \\ \hline
Baseline (MC $\rightarrow$ RedWeb $\rightarrow$ DIW ) & 1.313 & 0.1316 & 0.4551 \\ \hline
PI-only (No dataset supervision) & 1.299 & 0.1253 & 0.4491 \\
Ranking + PI (DIW) & 1.298 & 0.1202 & 0.4374 \\
Ranking + PI + PA (DIW + KD) & \textbf{1.294} & 0.1202 & 0.4396 \\
Ranking + PI + PA + GAN (DIW + KD) & \textbf{1.294} & 0.1204 & 0.4319 \\ \hline
Ranking + PI + PA  + 2 IP (DIW + KD) & 1.303 & \textbf{0.1198} & \textbf{0.4213} \\ \hline
\end{tabular}%
}
\medskip
\caption{Contributions of knowledge distillation approach. \textit{PI, PA and GAN} are pixelwise, pairwise and holistic losses \cite{liu2020structured}.} 
\vspace{-4mm} 
\label{tab:ablation_kd}
\end{table}
\endgroup

\textbf{Loss rebalancing}
We study the effect of loss balancing; we analyse initial loss weight scaling, emphasizing easy or hard losses first and wrapping auxiliary layers (i.e. not automatically rebalancing $\lambda$ values in Equation (\ref{eq:aux}), but fixing them instead). We perform loss rebalancing five times every epoch and base our experiments on the best model of previous section. We compare our approach with another loss balancing method \cite{malkiel2020mtadam}.  Results are shown in Table \ref{tab:ablation_lossrebalancing}.

Results show that loss rebalancing introduces further improvements. MTAdam \cite{malkiel2020mtadam} decreases our accuracy across the board. Loss weight initialization degrades our results severely, suggesting that balancing only the loss decay rate is better than balancing the loss value scales as well as the decay rate. Wrapping intermediate predictions as one loss term is the sensible choice; the alternative is to emphasize intermediate prediction losses, which means only early parts of the decoder will get the required supervisory signal. Lastly, we see learning hard losses and then easy losses works better than the alternative; when we focus on hard losses first, we emphasize pixelwise and pairwise losses initially, and then the ranking loss. Effectively, we perform the knowledge transfer first and then train our model with DIW ground-truth annotations. The rate of loss weight change is large for pairwise loss (0.33 to 0.1), pixelwise loss' weights are stable (0.33 to 0.285) but ranking loss gains attention in the later stages of the training (0.33 to 0.9).

\begingroup
\setlength{\tabcolsep}{8pt} % Default value: 6pt
\renewcommand{\arraystretch}{1.5} % Default value: 1
\begin{table}[]
\resizebox{\columnwidth}{!}{%
\begin{tabular}{lccc}
\hline
\textbf{Model and Training Data} & \textbf{RMSE $\downarrow$} & \textbf{WHDR $\downarrow$} & \textbf{SI-RMSE $\downarrow$} \\ \hline
Baseline (MC $\rightarrow$ RedWeb $\rightarrow$ DIW ) & 1.313 & 0.1316 & 0.4551 \\
Ours (no loss rebalancing) (DIW + KD) & 1.303 & 0.1198 & \textbf{0.4213} \\ \hline
Ours + MTAdam \cite{malkiel2020mtadam} (DIW + KD) & 1.316 & 0.1328 & 0.4704 \\
Ours + WI + easy $\rightarrow$ hard (DIW + KD) & 1.688 & 0.1332 & 0.5876 \\
Ours + no WI + easy $\rightarrow$ hard, FIP (DIW + KD) & 1.300 & 0.1226 & 0.4286 \\ \hline
Ours + no WI + hard $\rightarrow$ easy,  FIP (DIW + KD) & \textbf{1.300} & \textbf{0.1156} & 0.4292 \\ \hline
\end{tabular}%
}
\medskip
\caption{Contributions of multi-loss rebalancing approach. \textit{WI} and \textit{FIP} stand for loss weight initialization and fixed intermediate prediction weights, respectively. Arrows show which losses the algorithm considers the first (i.e. hard losses and then easy losses). } 
\vspace{-5mm} 
\label{tab:ablation_lossrebalancing}
\end{table}
\endgroup

\subsection{Final Results} \label{sec:final_results}

\begin{figure*}[!ht]%
\centering
\vspace{-6mm}
\includegraphics[width=\textwidth]{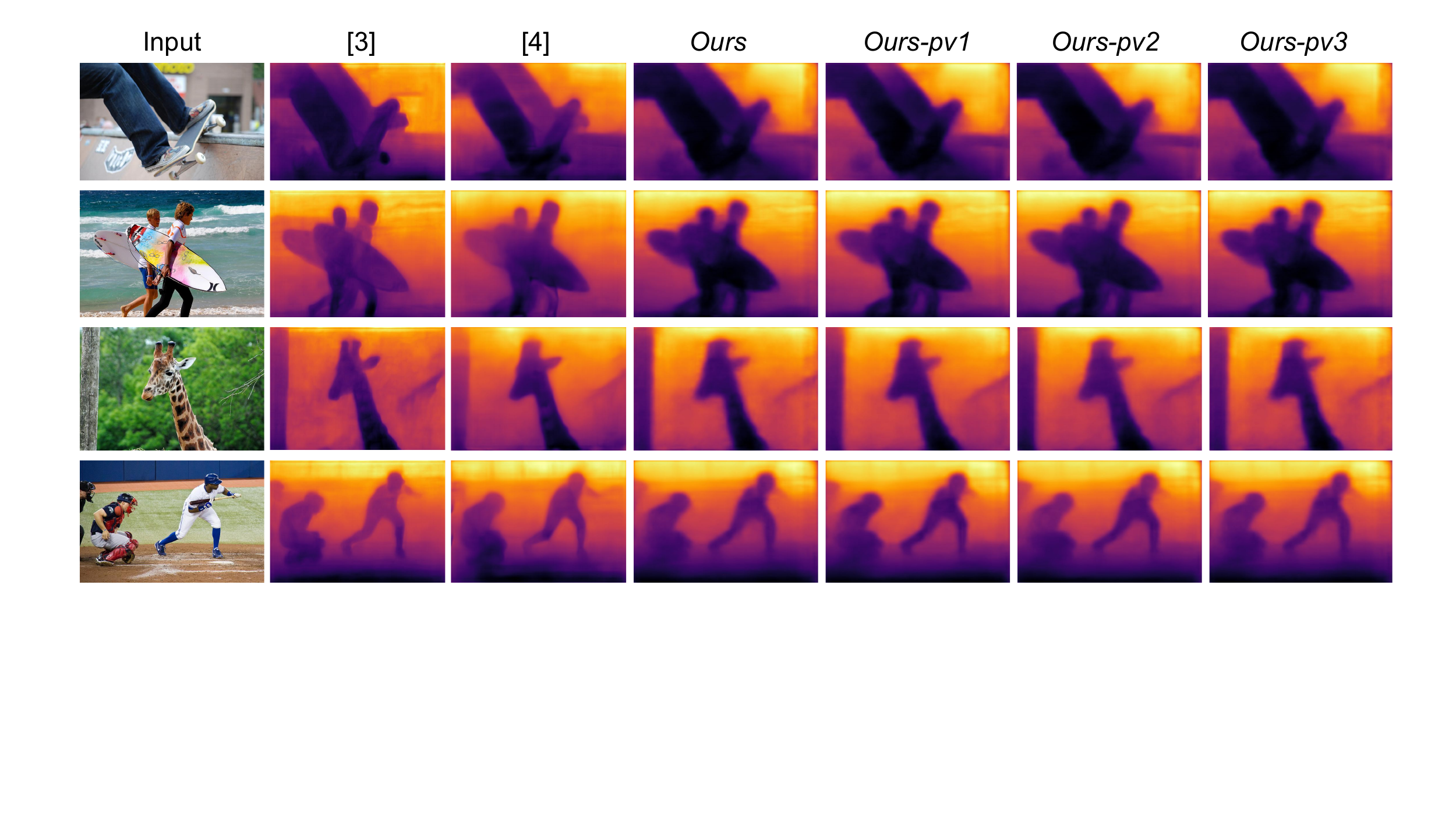}
\vspace{-3cm}
    \caption{Zero-shot qualitative results on COCO dataset. From left to right; input, \cite{chen2016single} (trained on NYUv2+DIW), \cite{chen2019learning} (trained on RedWeb+DIW+Y3D), \textit{ours}, \textit{ours-prunedv1}, \textit{ours-prunedv2} and \textit{ours-prunedv3}. \textit{Ours} are trained on DIW with knowledge distillation.} 
    
\label{fig:qualitative}
\vspace{-4mm}
\end{figure*}

We compare our model against state-of-the-art methods on DIW. We exclusively report WHDR scores as it is the primary evaluation metric for depth in-the-wild scenarios, as explained in Section \ref{sec:motivation}. \textit{Ours} is our best model (i.e. last row of Table \ref{tab:ablation_lossrebalancing}). Results are shown in Table \ref{tab:sota_comparison}.

Results show our approach has the best WHDR value where DIW is the only supervisory dataset. Models exceeding our results train on multiple datasets, which makes our model the most data efficient. Moreover, we train on a resolution of 224x448, whereas others train on larger resolutions (\cite{chen2019learning, mertan2020new, xian2018monocular,ranftl2019towards} train on 384x384)\footnote{ \cite{chen2016single} trains on 240x320. No such information is available for \cite{li2018megadepth}.}. A strong contender is \cite{ranftl2019towards}, where a lightweight model trained on 256x256 resolution achieves 0.1344 WHDR. This is an impressive result despite not being trained on DIW, however this model is trained on ten other datasets and ours is better by 0.2 WHDR despite being smaller. All the successful methods use ResNet \cite{he2016deep} architecture; our method outperforms them in cases where DIW is the only supervisory dataset,  with a significantly less complex model. We also report the results of the pruned versions of our model (last three rows of Table \ref{tab:sota_comparison}). Our pruned model is still the best where DIW is the only supervisory dataset, with only 1M parameters. 

Qualitative results are shown in Figure \ref{fig:qualitative}. Compared to larger models trained on multiple datasets including metric depth annotations, our results are competitive  despite being trained only on ordinal pairs. Runtime results are shown in Table \ref{tab:runtime}. \cite{chen2016single} \footnote{We use an unofficial implementation of \cite{chen2016single}, available at \url{https://github.com/Turmac/DIW_TF_Implementation}.} performs bad, especially on CPU, due to its non-mobile friendly architecture. \cite{ranftl2019towards} performs well, but can not achieve real-time on CPU. Compared with two other lightweight models, our model is significantly faster and reaches real-time on CPU and GPU with better WHDR. 

\textbf{Metric depth performance.}  We also report metric depth results despite training only on ordinal pairs. Two methods in Table \ref{tab:sota_comparison} report metric results on NYU with models trained only on ordinal pairs and none report SI-RMSE on MC. \cite{chen2016single} trains on NYU+DIW and \cite{xian2018monocular} trains on NYU and report 1.10 and 1.07 RMSE on NYU, respectively. Ours achieves 1.30 RMSE (last row of Table \ref{tab:ablation_lossrebalancing}), which is worse than the others. However, \cite{xian2018monocular} trains on NYU and thus "specialize" in indoor scenarios, and \cite{chen2016single} trains on significantly more data. Improved performance on metric depth using only relative annotations is a future venue we are planning to explore, but is not within the scope of this study.

\begingroup
\setlength{\tabcolsep}{8pt} % Default value: 6pt
\renewcommand{\arraystretch}{1.5} % Default value: 1
\begin{table}[t]
\resizebox{\columnwidth}{!}{%
\begin{tabular}{llcc}
\hline
\textbf{Method} & \textbf{Training Data} & \textbf{WHDR $\downarrow$} & \textbf{Parameter Count} \\ \hline
Chen \textit{et al.} \cite{chen2016single} & DIW & 0.2214 & 5.3M  \\
Chen \textit{et al.} \cite{chen2016single}& NYU + DIW & 0.1439 & 5.3M\\ \hline
Li \cite{li2018megadepth} & MegaDepth & 0.2297 &  5.3M \\  \hline
Xian \textit{et al.} \cite{xian2018monocular} & DIW & 0.1498 & 53.1M  \\
Xian \textit{et al.} \cite{xian2018monocular} & RW + DIW & 0.1137 &  53.1M\\ \hline
Ranftl \textit{et al.} \cite{ranftl2019towards} & RW + DL + MV + MD + WS & 0.1246 & 105.3M \\ 

Ranftl \textit{et al.} \cite{ranftl2019towards} & RW + DL + MV + MD + WS $\dagger$ & 0.1344 & 21.3M \\ \hline
Mertan \textit{et al.} \cite{mertan2020new} & DIW & 0.1250 & 53.1M \\
Mertan \textit{et al.} \cite{mertan2020new}  & RW + DIW & 0.1101 & 53.1M\\ \hline
Chen \textit{et al.} \cite{chen2019learning} & RW + DIW + Y3D & \textbf{0.1059} & 53.1M\\ \hline
Our baseline & MC-RW-DIW & 0.1316 & 2.6M \\
Ours & DIW (+ KD) & 0.1156 & 2.6M \\
Ours-pruned v1 & DIW (+ KD)  & 0.1190 & 1.9M \\
Ours-pruned v2 & DIW (+ KD)  & 0.1196 & 1.5M \\
Ours-pruned v3 & DIW (+ KD)  & 0.1208 & \textbf{1M} \\
 \hline
\end{tabular}%
}
\medskip
\caption{Comparison with the state-of-the-art baselines on DIW test set. Several methods use the same architecture, therefore parameter counts for methods with no released weights are assumed to be the same. Row denoted with $\dagger$ trains on 10 datasets. }
\vspace{-5mm} 
\label{tab:sota_comparison}
\end{table}
\endgroup

\begingroup
\setlength{\tabcolsep}{8pt} % Default value: 6pt
\renewcommand{\arraystretch}{1.5} % Default value: 1
\begin{table}[]
\resizebox{\columnwidth}{!}{%
\begin{tabular}{clccc}
\hline
\textbf{Device} & \textbf{Model} & \textbf{CPU} & \multicolumn{1}{l}{\textbf{GPU}} & \multicolumn{1}{l}{\textbf{NNAPI}} \\ \hline
\multirow{6}{*}{Samsung Galaxy S10+} & {\cite{ranftl2019towards}} & 274 & 99 & 332 \\
 & {\cite{chen2016single}} & \multicolumn{1}{l}{3361} & \multicolumn{1}{l}{390} & \multicolumn{1}{l}{\ 1790} \\ \cline{2-5} 
 & Ours & 116 & 74 & 108 \\
 & Ours -pruned v1 & 113 & 72 & 105 \\
 & Ours -pruned v2 & 104 & 69 & 103 \\
 & Ours -pruned v3 & 101 & 67 & 101 \\ \hline
\multirow{6}{*}{Samsung Galaxy S21} & {\cite{ranftl2019towards}} & 215 & 53 & 141 \\
 & {\cite{chen2016single}} & \multicolumn{1}{l}{3101} & \multicolumn{1}{l}{215} & \multicolumn{1}{l}{\  \ \ 375} \\ \cline{2-5} 
 & Ours & 54 & 27 & 43 \\
 & Ours -pruned v1 & 47 & 26 & 25 \\
 & Ours -pruned v2 & 43 & 25 & 35 \\
 & Ours -pruned v3 & 39 & 23 & 33 \\ \hline
\end{tabular}%
}
\medskip
\caption{Average runtime performances of our model and other lightweight methods \cite{ranftl2019towards,chen2016single}. Runtime values are in ms and measured with same input resolution (256x256).}
\vspace{-6mm} 
\label{tab:runtime}
\end{table}
\endgroup

\section{Conclusion} \label{conclude}

Monocular relative depth estimation is an important task in various applications, specifically in mobile settings. Therefore, it is imperative to design a model that is efficient and accurate in in-the-wild scenarios. To this end, we explore the design space of a lightweight relative depth estimator for mobile devices. Following a carefully performed model design process, we present a pipeline where we improve our model with a combination of knowledge distillation, intermediate predictions and loss rebalancing.   

Our model achieves real-time performance on both mobile CPU and GPU and reaches 0.1156 WHDR on DIW, the best result among models that use only DIW as the supervisory dataset, with a fraction of model complexity. Our model, trained only on relative depth annotations, without pruning or hardware optimizations, has 2.6M parameters and runs with 37 FPS on a mobile GPU. A significantly pruned version achieves 0.1208 WHDR on DIW with 1M parameters and runs with 44 FPS on a mobile GPU.

{\small
\bibliographystyle{ieee_fullname}
\bibliography{egbib}
}

\end{document}